\documentclass[journal]{IEEEtai}

\usepackage{color}
\usepackage{amsmath}
\usepackage{amssymb}
\usepackage{bm}
\usepackage{url}
\usepackage{graphicx}
\usepackage[colorlinks,urlcolor=blue,linkcolor=blue,citecolor=blue]{hyperref}
\usepackage{array}
\usepackage{multirow}
\usepackage{booktabs}
\usepackage{multirow}
\usepackage[figuresright]{rotating}
\usepackage{oubraces}
\usepackage{mathtools}
\usepackage{subfigure}
\usepackage{rotating}
\usepackage{tabu}
\usepackage[ruled,vlined]{algorithm2e}
\newcommand{\vc}[1]{\boldsymbol{\mathbf{#1}}}
\DeclareMathSymbol{\R}{\mathalpha}{AMSb}{"52}

\definecolor{bs}{rgb}{0.0, 0.0, 0.0}

\SetAlgoLined
\SetArgSty{textnormal}

\makeatletter
\newcommand{\removelatexerror}{\let\@latex@error\@gobble}
\makeatother

\begin{document}

\title{
FairDrop: Biased Edge Dropout for Enhancing Fairness in Graph Representation Learning}

\author{Indro Spinelli, Simone Scardapane, Amir Hussain, and Aurelio Uncini
%\thanks{Manuscript received xx xx, xx. (Corresponding author: Indro Spinelli.)}
\thanks{ I. Spinelli, S. Scardapane, and A. Uncini are with the Department of Information Engineering, Electronics and Telecommunications (DIET), Sapienza University of Rome,
00184 Rome, Italy (e-mail: indro.spinelli@uniroma1.it; simone.scardapane@uniroma1.it; aurelio.uncini@uniroma1.it)}
\thanks{A. Hussain is with the School of Computing, Edinburgh Napier University, UK (e-mail: a.hussain@napier.ac.uk).}}

\markboth{Preprint submitted to IEEE Transactions on Artificial Intelligence}
{Indro Spinelli \MakeLowercase{\textit{et al.}}: Fair Edge Dropout}

\maketitle

\begin{abstract}
Graph representation learning has become a ubiquitous component in many scenarios, ranging from social network analysis to energy forecasting in smart grids. In several applications, ensuring the fairness of the node (or graph) representations with respect to some protected attributes is crucial for their correct deployment. Yet, fairness in graph deep learning remains under-explored, with few solutions available. In particular, the tendency of similar nodes to cluster on several real-world graphs (i.e., homophily) can dramatically worsen the fairness of these procedures. In this paper, we propose a novel biased edge dropout algorithm (FairDrop) to counter-act homophily and improve fairness in graph representation learning. FairDrop can be plugged in easily on many existing algorithms, is efficient, adaptable, and can be combined with other fairness-inducing solutions. After describing the general algorithm, we demonstrate its application on two benchmark tasks, specifically, as a random walk model for producing node embeddings, and to a graph convolutional network for link prediction. We prove that the proposed algorithm can successfully improve the fairness of all models up to a small or negligible drop in accuracy, and compares favourably with existing state-of-the-art solutions. In an ablation study, we demonstrate that our algorithm can flexibly interpolate between biasing towards fairness and an unbiased edge dropout. Furthermore, to better evaluate the gains, we propose a new dyadic group definition to measure the bias of a link prediction task when paired with group-based fairness metrics. In particular, we extend the metric used to measure the bias in the node embeddings to take into account the graph structure.
\end{abstract}

\begin{IEEEImpStatement}
Fairness in graph representation learning is under-explored. Yet, the algorithms working with these types of data have a fundamental impact on our digital life. Therefore, despite the law now prohibiting unfair treatment by artificial intelligence (AI) methods (such as those based on sensitive traits, social networks and recommender systems) to systematically discriminate against minorities, current solutions are computationally intensive, deliver reduced performance accuracy and/or lack interpretability. To address the fairness problem, we propose FairDrop, a novel biased edge dropout algorithm. Our approach protects against unfairness generated from the network’s homophily with respect to sensitive attributes. FairDrop can be readily integrated into today’s AI solutions for learning network embeddings or downstream tasks. We believe the lack of expensive computations and flexibility of our proposed fairness constraint will posit FairDrop as a future benchmark resource that will serve to highlight and address fairness challenges for the global AI research and innovation community.
\end{IEEEImpStatement}

\begin{IEEEkeywords}
Graph representation learning, graph embedding, fairness, link prediction, graph neural network.
\end{IEEEkeywords}

\section{Introduction}
\label{sec:introduction}

\IEEEPARstart{G}{raph} structured data, ranging from friendships on social networks to physical links in energy grids, powers many algorithms governing our digital life. Social networks topologies define the stream of information we will receive, often influencing our opinion \cite{mcpherson2001homo}\cite{halberstam2016homo}\cite{Centola2010TheSO} \cite{lee2019fair} \cite{abbass2018social}. Bad actors, sometimes, define these topologies ad-hoc to spread false information \cite{roy2021fake}. Similarly, recommender systems \cite{ye2020rec} suggest products tailored to our own experiences and history of purchases. However, pursuing the highest accuracy as the only metric of interest has let many of these algorithms discriminate against minorities in the past \cite{datta2014ad} \cite{corbett2017fair} \cite{obermeyer2019}, despite the law prohibiting unfair treatment based on sensitive traits such as race, religion, and gender. For this reason, the research community has started looking into the biases introduced by machine learning algorithms working over graphs \cite{bose2019cfcge}.

One of the most common approaches to process graphs is via learning vector embeddings for the nodes (or the edges), e.g., \cite{grover2016n2v}. These are a low dimensional representation of the nodes (or the edges), encoding the local topology.  Downstream tasks then use these embeddings as inputs. Some examples of these tasks are node classification, community detection, and link prediction \cite{Cui2019ASO}. We will focus on the latter due to its widespread application in social networks and recommender systems. Alternatively, graph neural networks (GNNs) \cite{bronstein2017geometric} \cite{xiao2019saliency} solve link prediction or other downstream tasks in an end-to-end fashion, without prior learning of embeddings through ad-hoc procedures. The techniques developed in this paper can be applied to both scenarios.

In this work, we will concentrate on the bias introduced by one of the key aspects behind the success of GNNs and node embedding procedures: homophily. Homophily is the principle that similar users interact at a higher rate than dissimilar ones. In a graph, this means that nodes with similar characteristics are more likely to be connected. In node classification, this encourages smoothness in terms of label distributions and embeddings, yielding excellent results \cite{huang2021combining}.

From the fairness point of view, the homophily of sensitive attributes directly influences the prediction and introduces inequalities. In social networks, the ``unfair homophily" of race, gender or nationality, limits the contents accessible by the users, influencing their online behaviour \cite{mcpherson2001homo}. For example, the authors of \cite{halberstam2016homo} showed that users affiliated with majority political groups are exposed to new information faster and in greater quantity. Similarly, homophily can put minority groups at a disadvantage by restricting their ability to establish links with a majority group \cite{karimi2018homo}.
An unfair link prediction magnifies this issue, known as ``filter bubble",  by increasing the segregation between the groups. 

To mitigate this issue, in this paper we propose a biased dropout strategy that forces the graph topology to reduce the homophily of sensitive attributes. At each step of training, we compute a random copy of the adjacency matrix biased towards reducing the amount of homophily in the original graph. Thus, models trained on these modified graphs undergo a fairness-biased data augmentation regime. Our approach limits the biases introduced by the unfair homophily, resulting in fairer node representations and link predictions while preserving most of the original accuracy. While we focus on homophily, we underline that unfairness can arise from additional factors that we do not consider here (e.g., unfair weights in the adjacency matrix, or node feature vectors). In these cases, FairDrop can be easily combined with any other known fairness-inducing technique.

Measuring fairness in this context requires some adaptations. Most works on fairness measures focus on independent and identically distributed (i.i.d.) data. These works proposed many metrics, each one protecting against a different bias \cite{mehrabi2019ASO}. In particular, group fairness measures determine the level of equity of the algorithm predictions between groups of individuals. Link prediction requires a dyadic fairness measure that considers the influence of both sensitive attributes associated with the connection \cite{masrour2020fb}. However, it is still possible to apply group fairness metrics by defining new groups for the edges.

\subsection{Contributions of the paper}
We propose a preprocessing technique that modifies the training data to reduce the predictability of its sensitive attributes \cite{feldman2015preproc}. Our algorithm introduces no overhead and can be framed as a biased data augmentation technique. A single hyperparameter regulates the intensity of the fairness constraint. This ranges from maximum fairness to an unbiased edge dropout. Therefore FairDrop can be easily adapted to the needs of different datasets and models. These characteristics make our framework extremely adaptable. Our approach can also be applied in combination with other fairness constraints.
We evaluate the fairness improvement on two different tasks. Firstly, we measure the fairness imposed on the outcomes of end-to-end link prediction tasks. Secondly, we test the capability of removing the contributions of the sensitive attributes from the resulting node embeddings generated from representation learning algorithms.
To measure the improvements for the link prediction we also propose a novel group-based fairness metric on dyadic level groups. We define a new dyadic group and use it in combination with the ones described in \cite{masrour2020fb}.
Instead, for the predictability of the sensitive attributes from the node embeddings,  we measure the representation bias \cite{buyl2020debayes}. It is common to use node embeddings as input of downstream link prediction task.  Therefore, we introduce a new metric that also considers the graph's topology. 
\subsection{Outline of the paper}
The rest of the paper is structured as follows. Section \ref{sec:related_works} reviews recent works about enforcing fairness for graph-structured data and their limitations. Then, in Section \ref{sec:preliminaries} we introduce GNN models and group-based fairness metrics for i.i.d data. FairDrop, our proposed biased edge dropout, is first introduced in Section \ref{sec:method} and then tested in Section \ref{sec:evaluation}. We conclude with some general remarks in Section \ref{sec:conclusion}.

\section{Related Works}
\label{sec:related_works}
The literature on algorithmic bias is extensive and interdisciplinary \cite{romei2013ams}.
However, most approaches study independent and identically distributed data. Just recently, with the success of GNNs, some works started to investigate fairness in graph representation learning.
Some works focused on the creation of fair node embeddings \cite{bose2019cfcge}\cite{tahleen2019fairwalk} \cite{buyl2020debayes} that can be used as the input of a downstream task of link prediction. Others targeted directly the task of a fair link prediction \cite{masrour2020fb} \cite{li2021on}.

Some of these approaches base their foundations on adversarial learning. Compositional fairness constraints \cite{bose2019cfcge} learn a set of adversarial filters that remove information about particular sensitive attributes. Similarly, Fairness-aware link prediction \cite{masrour2020fb} employs an adversarial learning approach to ensure that inter-group links are well-represented among the predicted links.
We selected three of these contribution for our experimental evaluation.

\textbf{FairAdj} \cite{li2021on} learns a fair adjacency matrix during an end-to-end link prediction task. It exploits a graph variational autoencoder \cite{Kipf2016VariationalGA} and two different optimization processes, one for learning a fair version of the adjacency matrix and one for the link prediction.

\textbf{DeBayes} \cite{buyl2020debayes} extends directly ``Conditional Network Embedding" (CNE) \cite{kang2018conditional} to improve its fairness. CNE uses the Bayes rule to combine prior knowledge about the network with a probabilistic model for the Euclidean embedding conditioned on the network. Then, DeBayes maximizes the obtained posterior probability for the network conditioned on the embedding to yield a maximum likelihood embedding. DeBayes models the sensitive information as part of the prior distribution.

\textbf{Fairwalk} \cite{tahleen2019fairwalk} is an adaptation of Node2Vec \cite{grover2016n2v} that aims to increase the fairness of the resulting embeddings. It modifies the transition probability of the random walks at each step, by weighing the neighbourhood of each node, according to their sensitive attributes.

Outside of the graph domain, data augmentation strategies for ensuring fairness have been explored in the context of \textit{counterfactual fairness} \cite{kusner2017counterfactual,chiappa2019path} (CF). CF formalizes the idea that a certain protected attribute does not cause unfair decisions, by requiring the outcomes of a model to be equivalent in a counterfactual scenario where only the protected attribute is modified. However, CF-based data augmentation strategies require the complete specification of a casual model, making them difficult to apply in practice.

\section{Preliminaries}
\label{sec:preliminaries}
\subsection{Graph representation learning}

In this work we will consider an undirected and unweighted graph $\mathcal{G} = (\mathcal{V}, \mathcal{E})$, where $\mathcal{V} = \left\{1, \ldots, n\right\}$ is the set of node indexes, and $\mathcal{E} = \left\{(i, j) \; \vert \; i, j \in \mathcal{V}\right\}$ is the set of arcs (\textit{edges}) connecting pairs of nodes. The meaning of a single node or edge depends on the application. For some tasks a node $i$ is endowed with a vector $\vc{x}_i \in \mathbb{R}^d$ of features. Each node is also associated to a categorical sensitive attribute $s_i \in S$ (e.g., political preference, ethnicity, gender), which may or may not be part of its features. % $s_i \in [0, 1]$  in a bi-partite system
Connectivity in the graph can be summarized by the adjacency matrix $\vc{A} \in \left\{0, 1\right\}^{n\times n}$. This matrix is used to build different types of operators that define the communication protocols across the graph. The vanilla operator is the symmetrically normalized graph Laplacian \cite{kipf2017semi} $\widehat{\vc{L}} = \vc{D}^{-1/2}\vc{L}\vc{D}^{-1/2}$ with $\vc{D}$ being the diagonal degree matrix where $D_{ii} = \sum_j A_{ij}$, and $\vc{L}$ the Laplacian matrix $\vc{L} = \vc{D} - \vc{A}$.

We will focus on the task of link prediction where the objective is to predict whether two nodes in a network are likely to have a link \cite{nowell2007link}.
For graphs having vector of features associated to the nodes, these can combined with the structural information of the graph by solving an end-to-end optimization problem \cite{kipf2016variational}:
\begin{equation}
 \vc{\hat A} = \sigma(\vc Z\vc Z^T),  \text{with} \hspace{2px} \vc Z = \text{GNN}(\vc{X},\vc{A}) \,.
 \label{eq:linkpred}
\end{equation}
where $\text{GNN}(\cdot, \cdot)$ is a GNN operating on the graph structure (see \cite{bacciu2020gentle,zhang2020deep,wu2020comprehensive} for representative surveys on GNN models). Another approach first uses unsupervised learning techniques \cite{perozzi2014deepwalk} \cite{grover2016n2v} to create a latent representation of each node and then perform a downstream task of link prediction using the learned embeddings and standard classifiers (e.g., support vector machines). Differently from most GNNs, node embeddings are obtained in an unsupervised fashion, and they can be re-used for multiple downstream tasks. Recently, self-supervised models for GNNs have been explored to fill this gap \cite{jin2020self}. The framework we propose in this paper can be used to improve fairness in all these scenarios.

\subsection{Fairness measures in i.i.d. data}
\label{subsec:fairness_measures}

Fairness in decision making is broadly defined as the absence of any advantage or discrimination towards an individual or a group based on their traits \cite{nripsuta2019fdef}. This general definition leaves the door open to several different fairness metrics, each focused on a different type of discrimination \cite{mehrabi2019ASO}.
In this paper we focus on the definition of group fairness, known also as disparate impact. In this case, the unfairness occurs when the model's predictions disproportionately benefit or damage people of different groups defined by their sensitive attribute. 

These measures are usually expressed in the context of a binary classification problem. In the notation of the previous section, denote by $Y \in [0,1] $ a binary target variable, and by $\hat Y = f( \vc x)$ a predictor that does not exploit the graph structure. As before, we associate to each $\vc x$ a categorical sensitive attribute $S$. For simplicity's sake, we assume $S$ to be binary, but the following definitions extend easily to the multi-class case.
Two widely used criterion belonging to this group are \textit{demographic parity} \cite{dwork2012dp} and \textit{equalized odds} \cite{hardt2016eo}:
\begin{itemize}
    \item \textit{Demographic Parity} ($DP$) \cite{dwork2012dp}: $\hat Y$  satisfies $DP$ if the likelihood of a positive outcome is the same regardless of the value of the sensitive attribute $S$:
    \begin{equation}
        P(\hat Y|S = 1 ) = P(\hat Y|S = 0 )
        \label{eq:demographic_parity}
    \end{equation}
    \item \textit{Equalized Odds} ($EO$) \cite{hardt2016eo}:  $\hat Y$  satisfies $EO$ if it has equal rates for true positives and false positives between the two groups defined by the protected attribute $S$:
        \begin{equation}
        P(\hat Y = 1|S = 1 , Y = y ) = P(\hat Y = 1|S = 0, Y = y )
        \end{equation}
\end{itemize}
These definitions trivially extend to the case where the categorical sensitive attribute can have more than two values $|S| > 2$ and, for the rest of the paper, we will consider this scenario.
To measure the deviation from these ideal metrics it is common to use the differences between the groups defined by $S$. 
In particular the $DP$  difference is defined as the difference between the largest and the smallest group-level selection rate:
\begin{equation}
    \Delta DP = \max_s E[\hat Y | S = s] - \min_s E[\hat Y | S = s]
\end{equation}

For the $EO$ difference, we can report the largest discrepancy between the true positive rate difference and the false positive rate difference between the groups defined again by the protected attribute $S$ \cite{bird2020fairlearn}:
 \begin{align}  
    \Delta TPR & = \max_s E[\hat Y=1| S = s, Y=1] \\\nonumber
    & - \min_s E[\hat Y=1| S = s, Y= 1] \,,
\end{align}
 \begin{align}  
    \Delta FPR & = \max_s E[\hat Y=1| S = s, Y=0] \\\nonumber
    & - \min_s E[\hat Y=1| S = s, Y=0] \,.
\end{align}
The final metric is then given by:
\begin{equation}
\Delta EO = \max( \Delta TPR, \Delta FPR ) \,.
\end{equation}
However, these definitions are not suitable to be directly applied to graph-structured data.

\section{Proposed biased edge dropout to enhance fairness}
\label{sec:method}

\begin{figure*}
    \centering
    \includegraphics[width=0.95\textwidth]{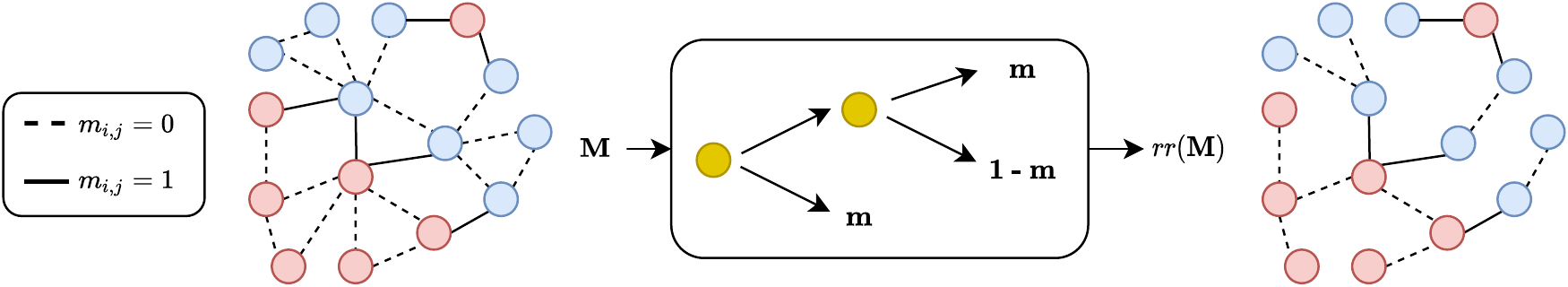}
    \caption{Schematics of FairDrop. From left to right: (i) Construction of the sensitive attribute homophily mask (Section \ref{sec:mask}); (ii) Randomized response used to inject randomness and to regulate the bias imposed by the constraint (Section \ref{sec:rr}); (iii) Final graph obtained by dropping the connections according to the randomized mask (Section \ref{sec:drop}).}
    \label{img:scheme}
\end{figure*}

In this work, for every epoch of training (either of the node embedding or the GNN), we generate a fairer random copy of the original adjacency matrix to counteract the negative effects introduced by the homophily of the sensitive attributes. To do so, we reduce the number of connections between nodes sharing the same sensitive attributes.

FairDrop can be divided into three main steps as shown in Figure \ref{img:scheme} and summarized in Algorithm \ref{alg:fairdrop}:
\begin{enumerate}
    \item Computation of a mask encoding the network's homophily with respect to the sensitive attributes.
    \item Biased random perturbation of the mask.
    \item Removing connections from the original adjacency matrix according to the computed perturbation.
\end{enumerate}

\begin{figure}[t]
\centering
\removelatexerror
\begin{algorithm}[H]
\SetKwInOut{Input}{Input}
\SetKwInOut{Output}{Output}
\Input{Adjacency matrix $\vc A$, sensitive attribute (vector) $\vc s$, mask for adjacency matrix initialized with zeros $\vc M$ , $\delta \in [0,\frac{1}{2}]$ \;}
\Output{$\vc A_{fair}$ Fair adjacency matrix }

 \ForAll{$\text{Edges} \left(i,j \right) \in \mathcal{G}$}{
   $m_{ij} = \vc{s}[i] \neq \vc{s}[j]$\;
    \eIf{$\textsc{random} \leq \frac{1}{2} + \delta$}{
    $rr(m_{ij}) = {m}_{ij}$ \;
   
   }
   {
   $rr(m_{ij}) = 1 - m_{ij}$\;
  
  }
  $\vc A_{fair} = \vc A \circ rr(\vc M)$\;
  
  }
 \caption{FairDrop}
\label{alg:fairdrop}
\end{algorithm}
\end{figure}

We conjecture that training a model over multiple fairness-enforcing random copies of the adjacency matrix (one for each epoch) increases the randomness and the diversity of its input. Thus the entire pipeline can be considered a data augmentation procedure biased towards fairness, similarly to how standard dropout can be interpreted as noise-based data augmentation \cite{helmbold2015inductive}.

\subsection{Sensitive attribute homophily mask}
\label{sec:mask}
We build an adjacency matrix $\vc M$ encoding the heterogeneous connections between the sensitive attributes. Its computation is the same for binary and categorical ones.
We define each element of $\vc M$ as:
\begin{equation}
m_{ij} = \begin{cases} 1 & \text { if } \:  s_i \neq s_j \\ 0 & \text{ otherwise} \end{cases} \,,
\end{equation}
where $m_{ij} = 1$  indicates an heterogeneous edge with respect to the sensitive attribute $S$. It is easy to see that if we want to improve the fairness of the graph we have to promote the first type of connections.

\subsection{Randomized response}
\label{sec:rr}
Randomized response \cite{sw1965rr} adds a layer of ``plausible deniability" for structured interviews. With this protection, respondents are able to answer sensitive issues while maintaining their privacy. The mechanism is easy and yet elegant. In its basic form, randomized response works with binary questions. The interviewee throws a coin to decide whether to respond honestly or at random with another coin toss. With unbiased coins, the user will tell the truth $2/3$ of the times and lie the remaining times.
More formally, randomized response can be defined as follows:
\begin{equation}
rr(m_{ij}) = \begin{cases} m_{ij} & \text { with probability}: \frac{1}{2} + \delta\\ 1-m_{ij} & \text { with probability}: \frac{1}{2} - \delta \end{cases} \,,
\end{equation}
with $\delta \in [0,\frac{1}{2}]$.
When $\delta = 0$ randomized response will always give random answers, maximizing the privacy protection but making the acquired data useless. Contrary, $\delta = 1/2$ gives us clean data without any privacy protection.

We use this framework, as shown in the central block of Fig. \ref{img:scheme}, where the input is the mask encoding the homophily of the sensitive attributes and the output will be its randomized version.

\subsection{Fair edge dropout}
\label{sec:drop}
Finally, we are ready to drop the unfair connection from the original adjacency matrix:
\begin{equation}
\vc A_{fair} = \vc A \circ rr(\vc M)
\end{equation}
where $\circ$ denotes the Hadamard product between matrices. The resulting matrix $\vc A_{fair}$ is a fairness-enforcing random copy of the original adjacency matrix. In particular, with $rr(\vc M)$ we bias the dropout towards fairness by cutting more connections between nodes sharing the same sensitive attribute. A model trained on this version of the graph will adapt consequently. The reduction of this unfair homophily will be reflected in the network embeddings or directly in the model's predictions.

It is easy to see that the parameter $\delta$ regulates the level of fairness we enforce on the topology. In fact, setting $\delta = 0$, corresponds to an unbiased edge dropout of $p= 1/2$ like the one proposed by \cite{rong2020dropedge}. Increasing $\delta$ will incrementally promote fair connections removing the unfair links in the graph. When $\delta$ is at its maximum value, FairDrop removes all the unfair edges and keeps all the fair ones.
The possibility of varying how much the algorithm is biased towards fairness makes it adaptable to the needs of different datasets and algorithms.
There are almost no additional costs from a standard edge dropout \cite{rong2020dropedge}, making this fairness ``constraint" extremely fast. Furthermore, FairDrop shares the capability of alleviating over-fitting and over-smoothing.

It is possible to apply FairDrop when multiple sensitive attributes are present at once. We can build a fair adjacency matrix for every sensitive attribute, and then these can be alternated for the training phase.

\subsection{Deployment}
FairDrop can be considered a data augmentation technique. It modifies the training data to reduce the influence of its sensitive attributes \cite{feldman2015preproc} \cite{calmon2017preproc}.
%This preprocessing can be paired with different models and in different learning scenarios.
We are interested in link prediction, both in an end-to-end fashion and by learning node embeddings and then predicting the links as a downstream task. For both approaches, we generate different random copies that enforce the fairness of the original graph. Like in \cite{rong2020dropedge} this augments the randomness and the diversity of the input data ingested by the learning model. Thus it can be considered a data augmentation biased towards fairness. Our approach can also be applied in combination with other fairness constraints.

In the first scenario, the model performs every training epoch on a different fair random copy of the input data. The model can be one of the many proposed in recent years that fit in the framework described by Equation \ref{eq:linkpred} like GCN \cite{kipf2017semi}, GAT \cite{velickovic2018graph}, AP-GCN \cite{spinelli2020apgcn} and many others. 

For learning the network embedding, we paired our edge sampling with Node2Vec \cite{grover2016n2v}. This algorithm learns a continuous representation for the nodes using local information obtained from truncated random walks \cite{perozzi2014deepwalk} balancing the exploration-exploitation trade-off. In this case, we generate fairness enforcing versions of the graph for every batch of random walks that we want to compute.

\section{Proposed fairness metrics for graph representation learning}

For graph-structured data, it is common to have attributes, including the sensitive ones, associated with the nodes. Therefore, if we are interested in evaluating the fairness of a node classification task, we can use the metrics described previously in Section \ref{subsec:fairness_measures}. However, for the link prediction and node embedding tasks, those metrics cannot be used straightforwardly. For link prediction, we need a dyadic fairness criterion expecting the predictions to be statistically independent from both sensitive attributes associated the edges. In Section \ref{subsec:fairness_link_prediction} we propose a novel set of fairness metrics for the task. For node embeddings, in Section \ref{subsec:fairness_node_embeddings} we extend the ideas presented in \cite{zemel2013rb} to the downstream task of link prediction.

\subsection{Fairness for link prediction}
\label{subsec:fairness_link_prediction}
\begin{figure*}
    \centering
    \includegraphics[width=0.95\textwidth]{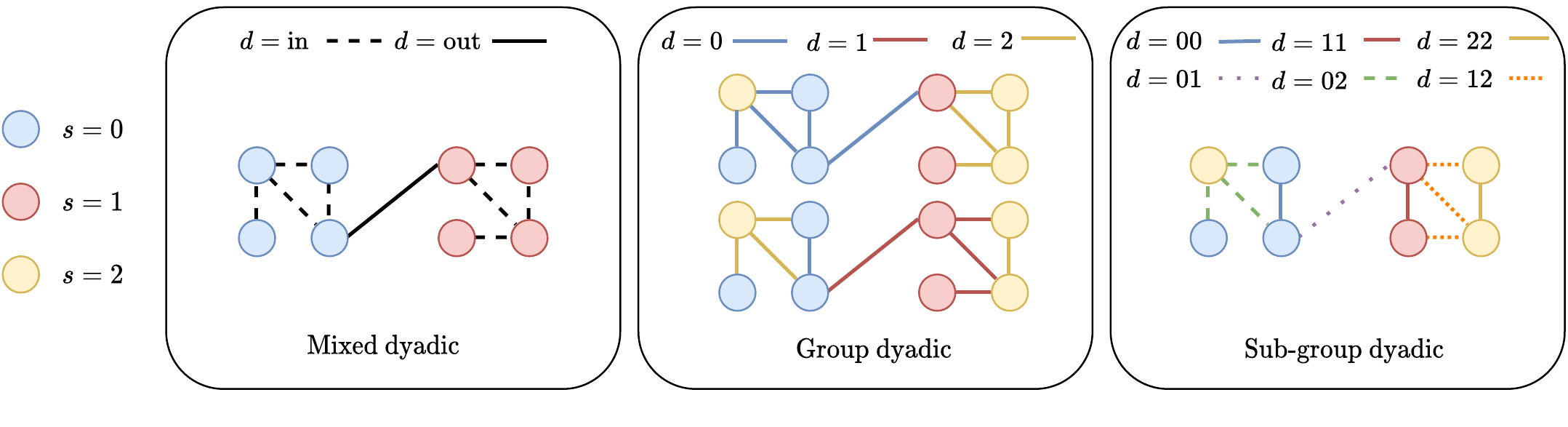}
    \caption{Visualization of the dyadic group $D$ definitions. We use a single sensitive attribute with $|S|= 3$. From left to right: (i) Mixed dyadic group definition generates two links group ``$d=\text{in}$" for intra connections and ``$d=\text{out}$" for inter ones; (ii) Group dyadic creates a one to one mapping between the sensitive attribute $S$ and the dyadic group $D$; (iii) The sub-group dyadic definition generates a group for every possible combination of the original sensitive attributes.}
    \label{img:groups}
\end{figure*}

\begin{figure*}
    \centering
    \includegraphics[width=0.95\textwidth]{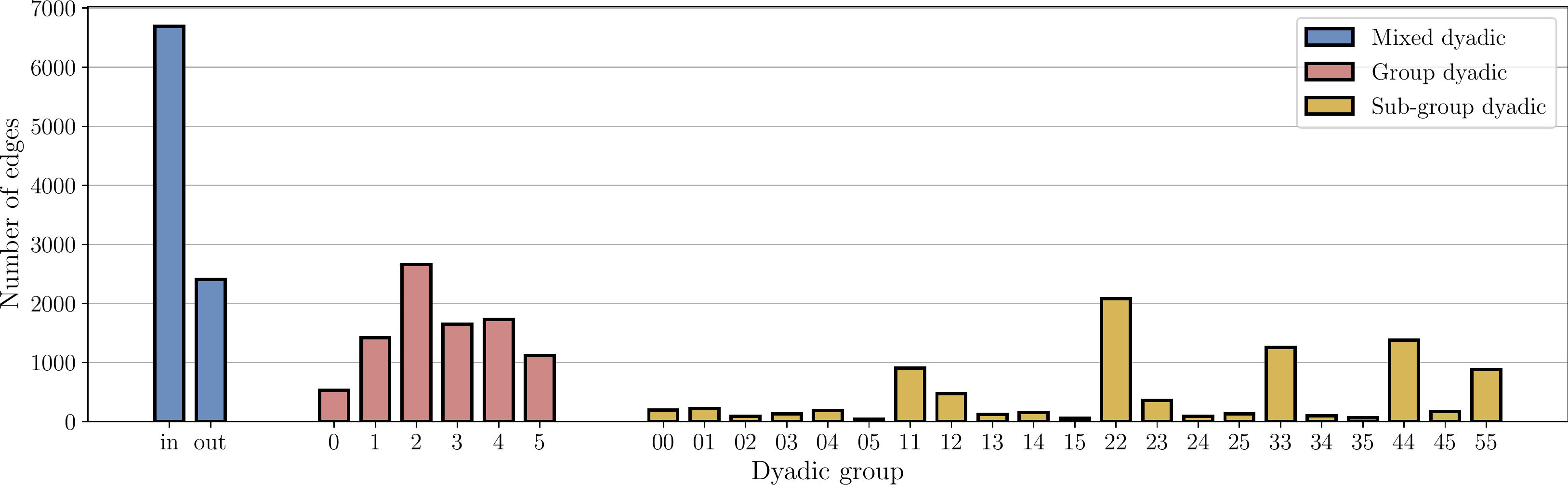}
    \caption{Barplot showing the distribution of the different dyadic groups for the Citeseer dataset.}
    \label{img:citeseer_groups}
\end{figure*}

An edge connects two nodes and thus two sensitive attributes. Our objective is to use the same fairness criteria for i.i.d. data ($DP$, $EO$) for the links. Therefore, we have to define so-called \textit{dyadic groups}, mapping the groups defined by the sensitive attribute for the nodes to new ones for the edges. It is thus possible to define different dyadic fairness criterion depending on how we build these dyadic groups.
In \cite{masrour2020fb} the authors introduced two dyadic criteria, mixed dyadic-level protection and subgroup dyadic-level protection. We propose a third criterion called group dyadic-level protection.  It aims to be closer to the original groups defined by the sensitive attributes. The original groups $S$ generate different dyadic subgroups associated with the edges $D$ of the graph instead of the nodes. In particular, from the coarsest to the finest, we have:
\begin{itemize}
    \item Mixed dyadic \cite{masrour2020fb} $\left(|D| = 2\right)$: the original groups determine two dyadic groups. An edge will be in the intra-group if it connects a pair of nodes with the same sensitive attribute. Otherwise, it will be part of the inter-group. This dyadic definition is common in recommender systems, where intra-group edges are known to create a ``filter bubble'' issue.
    
    \item Group dyadic $\left(|D| = |S|\right)$: there is a one-to-one mapping between the dyadic and node-level groups. To do so, we have to consider each edge twice, once for every sensitive attribute involved. A link is assigned to the group $D=s_1$ if it has one node with $S=s_1$, independent of the value of $S$ at the other end. The other end will determine the second assignment to $D=s_*$, where $*$ is the value of the sensitive attribute of the node not considered beforehand, of the same link. This dyadic definition ensures that the nodes participate in the links' creation regardless of the value of their sensitive attribute.
    
    \item Sub-group dyadic \cite{masrour2020fb} $\left(|D| = \frac{(|S|+2-1)!}{2!(|S|-1)!} \right)$: each dyadic sub-group represents a connection between two groups (or an intra-connection). The fairness criteria protect the balance between all the possible inter-group combinations together with the intra-group ones.
\end{itemize}
For example, considering the mixed dyadic group, and denoting the two groups as $D = 0$ for the inter-group and $D=1$ for the intra-group, we say that a link predictor achieves DP (extending \eqref{eq:demographic_parity}) if:
\begin{equation}
    P(\hat Y|D = 0) = P(\hat Y|D = 1 ) \,.
\end{equation}
All other metrics from Section \ref{subsec:fairness_measures} extend in a similar way for the three dyadic groups. We present a visualization of all the dyadic groups in Figure \ref{img:groups}. In Figure \ref{img:citeseer_groups} instead, we show the composition of these groups for the Citeseer dataset.

\subsection{Fairness for node embeddings}
\label{subsec:fairness_node_embeddings}
Representation learning aims to convert the graph's structure into a low-dimensional space. This representation is fair if it obfuscates as much as possible the unfair presence of sensitive attributes in the resulting embeddings \cite{zemel2013rb}.
Network embeddings are obfuscated if a classification task trying to predict the sensitive attribute yields the same accuracy as a random classifier \cite{bose2019cfcge} \cite{buyl2020debayes}. If this is the case, the embeddings are fair and the downstream task built upon them too. This type of bias afflicting node embeddings is also known as representation bias (RB). To measure the RB, we define an RB score \cite{buyl2020debayes} associated with the node embeddings $\vc Z$ as the balanced accuracy obtained by a classifier.
\begin{equation}
    \text{Node RB} = \sum_{s \in S} \frac{| \vc Z_s |}{|\mathcal{V}|} \text{Accuracy}\left(\vc Z_s \right)
\end{equation}
where $\vc Z_s$ is the set of node embeddings associated to nodes having $S=s$ and $\text{Accuracy}(\cdot)$ denotes the accuracy of a classifier trained to predict the sensitive attribute, restricted to the set $\vc Z_s$. Since we focus on link prediction, we also extend the definition of RB to the edges. We define the edge embeddings as the concatenation of the embeddings of the nodes participating in the links $\vc Z^{[src,dst]} $. We treat the classification problem as a multi-label since the edges have two sensitive attributes associated with their source nodes embedding $\vc Z^{src}$ and their destinations $\vc Z^{dst}$. We conjecture that the additional knowledge of the relation between the nodes helps the classification task. Consequently, it is harder to obtain a good score, making this metric more challenging. We define the representation bias score for the links as the mean of balanced accuracy obtained from the two target labels.
\begin{equation}
\begin{aligned}
    \text{Link RB} = \frac{1}{2}\sum_{s \in S} \frac{| \vc Z^{src}_s|}{|\mathcal{E}|} \text{Accuracy}_{src}\left(\vc Z^{[src,dst]}_s\right) \\ + \frac{|\vc Z^{dst}_s|}{|\mathcal{E}|} \text{Accuracy}_{dst}\left( \vc Z^{[src,dst]}_s\right) 
    \end{aligned}
\end{equation}
where $\text{Accuracy}_{src}(\cdot)$ is the accuracy with respect to the label of the source node, while $\text{Accuracy}_{dst}(\cdot)$ is the accuracy with respect to the label of the target node. We underline that the choice of a balanced metric for the representation bias may not be ideal when there is a strong under-representation of some groups. In these situations, an unbalanced version of the same metric can be preferred. It prevents the score from under-estimating a hypothetical perfect accuracy over an under-represented group. We return on this point in the conclusive section.

\section{Evaluation}
\label{sec:evaluation}
We test the fairness imposed by our algorithm by either evaluating the network embeddings obtained with Node2Vec, and on a downstream task of link prediction. We summarize the datasets used in this evaluation in Table \ref{tab:datasets}. We performed each experiment 10 times with different random seeds and data splits (80\%/20\% train and test) and we report the average with standard deviations. For the entire evaluation, we kept $\delta$ fixed to an intermediate value of 0.25.

\begin{table}[t]
\centering
\caption{Dataset statistics.}
\begin{tabular}{l|c|c|c|c|c}
Dataset & $S$ & $|S|$ & Features & Nodes & Edges\\
\hline
Citeseer & paper class & 6 & 3703 & 2110 & 3668 \\
Cora-ML  & paper class & 7 & 2879 & 2810 & 7981 \\
PubMed  & paper class & 3 & 500 & 19717 & 44324 \\
DBLP  & continent & 5 & None & 3980 & 6965 \\
FB & gender & 2 & None & 4039 & 88234
\end{tabular}
\label{tab:datasets}
\end{table}

\subsection{Fair link prediction}
\label{subsec:link_pred}
In these experiments, we test FairDrop's effect on an end-to-end link prediction task.
We tested a GCN \cite{kipf2017semi} and a GAT \cite{velickovic2018graph}, both with two layers, hidden dimension of 128 and trained with Adam optimizer \cite{kingma2014adam} for 200 epochs with a learning rate of 0.005. We performed the optimization once on the original adjacency matrix, then with unbiased edge dropout ($p=0.5$) and finally with our fairness enforcing dropout. We included in the evaluation FairAdj in both configurations used in the original paper.
We use the three benchmark citation networks from Table \ref{tab:datasets}, i.e., Citeseer, Cora-ML, and PubMed. In these graphs, nodes are articles. They have associated a bag-of-words representation of the abstract. Links represent a citation regardless of the direction.
The sensitive attribute, in this case, is the category of the article. We measured the link prediction quality with accuracy and the area under the ROC curve (AUC), the latter representing the trade-off between true and false positives for different thresholds. To evaluate the fairness of the predictions, we used the demographic parity difference ($\Delta DP$) and the equality of odds difference ($\Delta EO$) using the dyadic groups described in Section \ref{sec:introduction}. We report the results in Tables \ref{tab:cora}, \ref{tab:citeseer} and \ref{tab:pubmed} where the subscript of the fairness metrics is the initial of the dyadic group over which we computed the quantity. Arrows in the first row describes whether higher or lower values are better for the respective metric.

\begin{table*}[t]
\centering
\caption{Link prediction on Cora}
\begin{tabular}{l|c|c|c c|c c |c c}
 Method & Accuracy $\uparrow$ & AUC $\uparrow$ &  $\Delta DP_{m} \downarrow$ & $\Delta EO_{m}  \downarrow$ & $\Delta DP_{g}  \downarrow$ & $\Delta EO_{g}  \downarrow$ & $\Delta DP_{s}  \downarrow$ & $\Delta EO_{s}  \downarrow$\\
\midrule
GCN & 81.0 \scriptsize$\pm$ 1.1 & 88.0 \scriptsize$\pm$ 1.0 & 53.5 \scriptsize$\pm$ 2.4 & 34.8 \scriptsize$\pm$ 5.0 & 13.6 \scriptsize$\pm$ 3.2 & 17.7 \scriptsize$\pm$ 4.1 & 88.3 \scriptsize$\pm$ 3.3 & 100.0 \scriptsize$\pm$ 0.0  \\
GAT & 80.2 \scriptsize$\pm$ 1.4 & 88.3 \scriptsize$\pm$ 1.1 & 54.9 \scriptsize$\pm$ 2.9 & 39.6 \scriptsize$\pm$ 4.1 & 12.2 \scriptsize$\pm$ 2.5 & 16.5 \scriptsize$\pm$ 3.4 & 90.9 \scriptsize$\pm$ 3.5 & 100.0 \scriptsize$\pm$ 0.0 \\
\midrule
GCN+EdgeDrop & \textbf{82.4} \scriptsize$\pm$ 0.9 & \textbf{90.1} \scriptsize$\pm$ 0.7 & 56.4 \scriptsize$\pm$ 2.4 & 36.5 \scriptsize$\pm$ 4.3 & 12.3 \scriptsize$\pm$ 2.6 & 15.4 \scriptsize$\pm$ 3.3 & 90.2 \scriptsize$\pm$ 2.7 & 100.0 \scriptsize$\pm$ 0.0 \\
GAT+EdgeDrop & 80.5 \scriptsize$\pm$ 1.2 & 88.3 \scriptsize$\pm$ 0.8 & 53.7 \scriptsize$\pm$ 2.5 & 37.1 \scriptsize$\pm$ 3.2 & 18.8 \scriptsize$\pm$ 3.6 & 22.5 \scriptsize$\pm$ 4.2 & 93.6 \scriptsize$\pm$ 2.9 & 100.0 \scriptsize$\pm$ 0.0 \\

\midrule
FairAdj$_{T2=5}$ & 75.9 \scriptsize$\pm$ 1.6 & 83.0 \scriptsize$\pm$ 2.2 & 40.7 \scriptsize$\pm$ 4.1 & 20.9 \scriptsize$\pm$ 4.3 & 18.4 \scriptsize$\pm$ 2.8 & 31.9 \scriptsize$\pm$ 7.06 & 83.8 \scriptsize$\pm$ 4.9 & \textbf{98.3} \scriptsize$\pm$ 7.2 \\
FairAdj$_{T2=20}$ & 71.8 \scriptsize$\pm$ 1.6 & 79.0 \scriptsize$\pm$ 1.9 & \textbf{32.3} \scriptsize$\pm$ 2.8 & \textbf{15.8} \scriptsize$\pm$ 4.3 & 23.0 \scriptsize$\pm$ 4.2 & 41.4 \scriptsize$\pm$ 5.9 & \textbf{78.34} \scriptsize$\pm$ 6.8 & \textbf{98.3} \scriptsize$\pm$ 7.2 \\
\midrule
GCN+FairDrop & \textbf{82.4} \scriptsize$\pm$ 0.9 & \textbf{90.1} \scriptsize$\pm$ 0.7 & 52.9 \scriptsize$\pm$ 2.5 & 31.0 \scriptsize$\pm$ 4.9 & \textbf{11.8} \scriptsize$\pm$ 3.2 & \textbf{14.9} \scriptsize$\pm$ 3.7 & 89.4 \scriptsize$\pm$ 3.4 & 100.0 \scriptsize$\pm$ 0.0 \\
GAT+FairDrop & 79.2 \scriptsize$\pm$ 1.2 & 87.8 \scriptsize$\pm$ 1.0 & 48.9 \scriptsize$\pm$ 2.8 & 31.9 \scriptsize$\pm$ 4.3 & 15.3 \scriptsize$\pm$ 3.2 & 18.1 \scriptsize$\pm$ 3.5 & 94.5 \scriptsize$\pm$ 2.0 & 100.0 \scriptsize$\pm$ 0.0 \\
\end{tabular}
\label{tab:cora}
\end{table*}

\begin{table*}[t]
\centering
\caption{Link prediction on Citeseer}
\begin{tabular}{l|c|c|c c|c c |c c}
 Method & Accuracy $\uparrow$ & AUC $\uparrow$ &  $\Delta DP_{m} \downarrow$ & $\Delta EO_{m}  \downarrow$ & $\Delta DP_{g}  \downarrow$ & $\Delta EO_{g}  \downarrow$ & $\Delta DP_{s}  \downarrow$ & $\Delta EO_{s}  \downarrow$\\
\midrule

GCN & 76.7 \scriptsize$\pm$ 1.3 & 86.7 \scriptsize$\pm$ 1.3 & 42.6 \scriptsize$\pm$ 3.7 & 27.9 \scriptsize$\pm$ 4.7 & 20.6 \scriptsize$\pm$ 4.1 & 22.2 \scriptsize$\pm$ 4.6 & 68.1 \scriptsize$\pm$ 3.7 & 71.4 \scriptsize$\pm$ 9.1 \\
GAT & 76.3 \scriptsize$\pm$ 1.4 & 85.6 \scriptsize$\pm$ 1.9 & 42.4 \scriptsize$\pm$ 2.8 & 26.4 \scriptsize$\pm$ 4.1 & 21.1 \scriptsize$\pm$ 3.8 & 25.4 \scriptsize$\pm$ 5.6 & 71.3 \scriptsize$\pm$ 5.7 & 73.4 \scriptsize$\pm$ 9.9 \\
\midrule
GCN+EdgeDrop & 78.9 \scriptsize$\pm$ 1.3 & 88.0 \scriptsize$\pm$ 1.3 & 44.9 \scriptsize$\pm$ 2.5 & 27.5 \scriptsize$\pm$ 4.1 & 20.1 \scriptsize$\pm$ 2.9 & 21.6 \scriptsize$\pm$ 5.0 & 71.0 \scriptsize$\pm$ 3.4 & 73.2 \scriptsize$\pm$ 9.5 \\
GAT+EdgeDrop & 76.3 \scriptsize$\pm$ 0.9 & 85.6 \scriptsize$\pm$ 1.0 & 42.6 \scriptsize$\pm$ 2.5 & 28.4 \scriptsize$\pm$ 5.0 & 22.2 \scriptsize$\pm$ 5.1 & 27.6 \scriptsize$\pm$ 6.3 & 76.7 \scriptsize$\pm$ 3.0 & 77.5 \scriptsize$\pm$ 8.8 \\
\midrule
FairAdj$_{T2=5}$ & 78.5 \scriptsize$\pm$ 2.2 & 86.7 \scriptsize$\pm$ 2.2 & 39.2 \scriptsize$\pm$ 3.2 & 19.0 \scriptsize$\pm$ 3.9 & 17.3 \scriptsize$\pm$ 4.4 & 18.2 \scriptsize$\pm$ 5.8 & 62.6 \scriptsize$\pm$ 4.1 & 47.6 \scriptsize$\pm$ 8.8 \\
FairAdj$_{T2=20}$ & 74.4 \scriptsize$\pm$ 2.5 & 82.5 \scriptsize$\pm$ 2.7 & \textbf{31.0} \scriptsize$\pm$ 3.1 & \textbf{15.6} \scriptsize$\pm$ 3.0 & \textbf{8.8} \scriptsize$\pm$ 3.2 & 19.7 \scriptsize$\pm$ 6.9 & \textbf{56.1} \scriptsize$\pm$ 3.8 & \textbf{43.1} \scriptsize$\pm$ 7.4 \\
\midrule
GCN+FairDrop & \textbf{79.2} \scriptsize$\pm$ 1.4 & \textbf{88.4} \scriptsize$\pm$ 1.4 & 42.6 \scriptsize$\pm$ 2.5 & 26.5 \scriptsize$\pm$ 4.2 & 18.7 \scriptsize$\pm$ 4.0 & \textbf{17.6} \scriptsize$\pm$ 5.5 & 67.7 \scriptsize$\pm$ 3.5 & 64.3 \scriptsize$\pm$ 9.5 \\
GAT+FairDrop & 78.2 \scriptsize$\pm$ 1.1 & 87.1 \scriptsize$\pm$ 1.1 & 42.9 \scriptsize$\pm$ 2.2 & 28.3 \scriptsize$\pm$ 4.3 & 22.5 \scriptsize$\pm$ 3.4 & 25.9 \scriptsize$\pm$ 5.2 & 75.3 \scriptsize$\pm$ 3.2 & 73.4 \scriptsize$\pm$ 9.1 \\
\end{tabular}
\label{tab:citeseer}
\end{table*}

\begin{table*}[t]
\centering
\caption{Link prediction on PubMed}
\begin{tabular}{l|c|c|c c|c c |c c}
 Method & Accuracy $\uparrow$ & AUC $\uparrow$ &  $\Delta DP_{m} \downarrow$ & $\Delta EO_{m}  \downarrow$ & $\Delta DP_{g}  \downarrow$ & $\Delta EO_{g}  \downarrow$ & $\Delta DP_{s}  \downarrow$ & $\Delta EO_{s}  \downarrow$\\
\midrule
GCN & 88.0 \scriptsize$\pm$ 0.4 & 94.5 \scriptsize$\pm$ 0.2 & 43.9 \scriptsize$\pm$ 1.2 & 13.2 \scriptsize$\pm$ 1.4 & 5.0 \scriptsize$\pm$ 1.7 & \textbf{4.9} \scriptsize$\pm$ 1.7 & 57.3 \scriptsize $\pm$ 2.0 & 26.2 \scriptsize$\pm$ 3.6 \\
GAT & 80.8 \scriptsize$\pm$ 0.4 & 89.4 \scriptsize$\pm$ 0.3 & 42.3 \scriptsize$\pm$ 1.7 & 23.2 \scriptsize$\pm$ 1.9 & 2.3 \scriptsize$\pm$ 1.2 & 5.3 \scriptsize$\pm$ 1.2 & 59.0 \scriptsize$\pm$ 1.7 & 49.7 \scriptsize$\pm$ 3.4 \\
\midrule
GCN+EdgeDrop & 88.0 \scriptsize$\pm$ 0.5 & 94.6 \scriptsize$\pm$ 0.3 & 43.7 \scriptsize$\pm$ 1.0 & 12.8 \scriptsize$\pm$ 0.8 & 6.3 \scriptsize$\pm$ 0.7 & 6.0 \scriptsize$\pm$ 1.1 & 57.5 \scriptsize$\pm$ 1.4 & \textbf{26.3} \scriptsize$\pm$ 2.3 \\
GAT+EdgeDrop & 80.6 \scriptsize$\pm$ 0.9 & 88.8 \scriptsize$\pm$ 0.7 & 43.5 \scriptsize$\pm$ 1.1 & 24.5 \scriptsize$\pm$ 1.9 & 4.8 \scriptsize$\pm$ 1.6 & 7.5 \scriptsize$\pm$ 1.5 & 60.1 \scriptsize$\pm$ 1.9 & 49.3 \scriptsize$\pm$ 3.6 \\
\midrule
FairAdj$_{T2=5}$ & 75.5 \scriptsize$\pm$ 2.5 & 84.1 \scriptsize$\pm$ 2.2 & 32.3 \scriptsize$\pm$ 4.7 & 15.9 \scriptsize$\pm$ 4.7 & 7.3 \scriptsize$\pm$ 3.0 & 13.8 \scriptsize$\pm$ 6.2 & 53.4 \scriptsize$\pm$ 9.9 & 43.2 \scriptsize$\pm$ 9.5 \\
FairAdj$_{T2=20}$ & 73.8 \scriptsize$\pm$ 2.4 & 82.1 \scriptsize$\pm$ 2.0 & \textbf{28.9} \scriptsize$\pm$ 4.2 & 14.0 \scriptsize$\pm$ 4.0 & 7.8 \scriptsize$\pm$ 4.0 & 16.5 \scriptsize$\pm$ 6.7 & \textbf{52.5} \scriptsize$\pm$ 9.7 & 43.5 \scriptsize$\pm$ 9.8 \\
\midrule
GCN+FairDrop & \textbf{88.4} \scriptsize$\pm$ 0.4 & \textbf{94.8} \scriptsize$\pm$ 0.2 & 42.5 \scriptsize$\pm$ 0.5 & \textbf{12.2} \scriptsize$\pm$ 0.7 & 5.6 \scriptsize$\pm$ 1.8 & 5.1 \scriptsize$\pm$ 0.9 & 55.7 \scriptsize$\pm$ 1.5 & 26.6 \scriptsize$\pm$ 2.6 \\
GAT+FairDrop & 79.0 \scriptsize$\pm$ 0.8 & 87.6 \scriptsize$\pm$ 0.7 & 37.4 \scriptsize$\pm$ 0.9 & 19.7 \scriptsize$\pm$ 1.1 & \textbf{2.0} \scriptsize$\pm$ 1.0 & 6.4 \scriptsize$\pm$ 1.4 & 56.8 \scriptsize$\pm$ 2.1 & 47.3 \scriptsize$\pm$ 4.1 \\
\end{tabular}
\label{tab:pubmed}
\end{table*}

The models trained with FairDrop achieved superior performances in the link prediction tasks. FairDrop improved the fairness metrics of every dataset over an unbiased dropout or no dropout at all. Finally, FairDrop performances are close to FairAdj and on some occasions even better. \textcolor{bs}{However, FairAdj is far more complex and computationally expensive than our proposed solution. It takes 16 and 41 seconds, respectively, to train a fair GCN on Cora with FairAdj with $T2=5$ and $T2=20$. FairDrop requires just 2 seconds.}

\textcolor{bs}{As a general observation, the sub-group dyadic metrics were especially challenging to minimize: Cora in particular presented very unfair predictions with every algorithm taken into consideration. We believe that these results are due to the characteristics of the dataset itself. Cora shows the highest cardinality of the sensitive attribute. Furthermore, it has the largest imbalance between the ratios of inter-group and intra-group connections with the latter being dominant. We leave the improvement on these metrics to future work.}

\subsection{Fair representation learning}

Next, we compare our fairness constraint against two state-of-the-art frameworks for debiasing node embeddings, DeBayes \cite{buyl2020debayes} and FairWalk \cite{tahleen2019fairwalk}. We compute both node and link representation bias scores described in Section \ref{sec:introduction} and the classification accuracy on the downstream task of link prediction, using three different families of classifiers. In particular, we compare a logistic regression (LR), a three-layer neural network (NN) and a random forest (RF). To compute the optimal hyperparameters for each task and each one of these classifiers, we perform a grid search with cross-validation. We repeat this step for each one of the generated embeddings to be as fair as possible in the evaluation.
Ideally, the best algorithm will achieve the lowest scores on both representation biases while keeping the accuracy high.

For this task we use a co-authorship network built in \cite{buyl2020debayes} from DBLP \cite{tang2008dbpl}. They extracted the authors from the accepted papers of eight different conferences. The nodes of the graphs representing the authors are linked if they have collaborated at least once. The sensitive attribute is the continent of the author institution. The continents of Africa and Antarctica are not considered due to their under-representation in the data. The other dataset we considered is FB \cite{leskovec2012data} which is a combination of ego-networks obtained from a social network.  The graph encodes users as nodes with gender as a sensitive attribute and friendships as links. 

We use the same hyperparameters of Node2Vec for FairWalk and FairDrop: 50 training epochs, Adam optimizer with a learning rate of 0.01, 10 random walks of length 30 for each node in the graph. For DeBayes, we used the parameters provided in their implementation.

We present the results in Tables \ref{tab:dblp} and \ref{tab:fb} respectively for DBLP and FB datasets, and for the dataset used also in Section \ref{subsec:link_pred}, in Tables \ref{tab:cora_emb}, \ref{tab:citeseer_emb} and \ref{tab:pubmed_emb}.

\begin{table}
	\caption{Representation Learning on DBLP Dataset}
	%\fontsize{9pt}
	%\setlength{\arraystretch}{0.5}
	\linespread{1.0} 
	\setlength{\tabcolsep}{3pt} % Default value: 6pt
	\centering
	\begin{tabu}{l|p{11mm}|p{13mm}|p{13mm}|p{13mm}}
	
	Method & Classifier & Node RB $\downarrow$ & Link RB $\downarrow$ & Accuracy $\uparrow$ \\
\midrule
DeBayes & LR \newline NN \newline RF & \textbf{20.0} {\scriptsize$\pm$ 0.0} \newline  63.9 {\scriptsize$\pm$ 4.3}\newline 20.9 {\scriptsize$\pm$ 0.3} & \textbf{21.6} {\scriptsize$\pm$ 1.0}\newline 77.4 {\scriptsize$\pm$ 3.4 }\newline 28.0 {\scriptsize$\pm$ 0.6} & 87.2 {\scriptsize$\pm$ 0.3}\newline 91.2 {\scriptsize$\pm$ 0.7}\newline \textbf{91.6} {\scriptsize$\pm$ 0.5}\\
%\midrule
%Node2Vec  & LR \newline NN \newline RF & 52.0 {\scriptsize$\pm$ 1.2}\newline  67.5 {\scriptsize$\pm$ 1.0}\newline 53.7 {\scriptsize$\pm$ 0.4} & 65.8 {\scriptsize$\pm$ 1.2 }\newline 85.6 {\scriptsize$\pm$ 1.7 }\newline 70.9 {\scriptsize$\pm$ 0.2} & 90.3 {\scriptsize$\pm$ 0.6}\newline 87.7 {\scriptsize$\pm$ 1.1}\newline 88.9 {\scriptsize$\pm$ 0.5}\\
\midrule
FairWalk & LR \newline NN \newline RF & 53.5 {\scriptsize$\pm$ 5.3}\newline  74.2 {\scriptsize$\pm$ 4.9}\newline 20.8 {\scriptsize$\pm$ 0.7} & 67.1 {\scriptsize$\pm$ 3.4 }\newline 81.9 {\scriptsize$\pm$ 3.0}\newline 28.2 {\scriptsize$\pm$ 0.4} & \textbf{92.9} {\scriptsize$\pm$ 0.6}\newline \textbf{93.5} {\scriptsize$\pm$ 0.9}\newline 90.9 {\scriptsize$\pm$ 0.7}\\
\midrule
Node2Vec+FairDrop & LR \newline NN \newline RF & 31.6 {\scriptsize$\pm$ 3.5}\newline  \textbf{34.4} {\scriptsize$\pm$ 3.5}\newline \textbf{20.0} {\scriptsize$\pm$ 0.0} & 58.8 {\scriptsize$\pm$ 4.3}\newline \textbf{71.7} {\scriptsize$\pm$ 3.1}\newline \textbf{26.4} {\scriptsize$\pm$ 0.3} & 88.4 {\scriptsize$\pm$ 0.8}\newline 88.1 {\scriptsize$\pm$ 0.6}\newline 87.2 {\scriptsize$\pm$ 0.7}\\
	\end{tabu}
	\label{tab:dblp}
\end{table}

\begin{table}
    \caption{Representation Learning on FB Dataset}
	%\fontsize{9pt}
	%\setlength{\arraystretch}{0.5}
	\linespread{1.0} 
	\setlength{\tabcolsep}{3pt} % Default value: 6pt
	\centering
	\begin{tabu}{l|p{11mm}|p{13mm}|p{13mm}|p{13mm}}
	
	Method & Classifier & Node RB $\downarrow$ & Link RB $\downarrow$ & Accuracy $\uparrow$\\
    \midrule
    DeBayes & LR \newline NN \newline RF & 52.0 {\scriptsize$\pm$ 0.5} \newline  58.9 {\scriptsize$\pm$ 1.3}\newline 53.4 {\scriptsize$\pm$ 0.6}  & \textbf{55.7} {\scriptsize$\pm$ 0.5}\newline \textbf{91.6} {\scriptsize$\pm$ 0.5}\newline 62.7 {\scriptsize$\pm$ 0.6} & 96.7 {\scriptsize$\pm$ 0.1}\newline \textbf{98.5} {\scriptsize$\pm$ 0.2} \newline \textbf{97.9} {\scriptsize$\pm$ 0.2}\\
    %\midrule
    %Node2Vec  & LR \newline NN \newline RF & 61.6 {\scriptsize$\pm$ 0.8}\newline  58.0 {\scriptsize$\pm$ 1.3}\newline 62.3 {\scriptsize$\pm$ 0.7} & 68.1 {\scriptsize$\pm$ 0.7}\newline 98.7 {\scriptsize$\pm$ 0.3}\newline 90.8 {\scriptsize$\pm$ 0.3} & 96.4 {\scriptsize$\pm$ 0.1}\newline 96.5 {\scriptsize$\pm$ 0.3}\newline 96.4 {\scriptsize$\pm$ 0.1}\\
    \midrule
    FairWalk & LR \newline NN \newline RF & 53.1 {\scriptsize$\pm$ 0.6}\newline  58.9 {\scriptsize$\pm$ 1.3}\newline 53.5 {\scriptsize$\pm$ 0.6} & 63.3 {\scriptsize$\pm$ 1.0} \newline 99.5 {\scriptsize$\pm$ 0.2} \newline 62.3 {\scriptsize$\pm$ 0.8} & \textbf{97.4} {\scriptsize$\pm$ 0.2} \newline 97.6 {\scriptsize$\pm$ 0.3}\newline 97.0 {\scriptsize$\pm$ 0.2}\\
    \midrule
    Node2Vec+FairDrop & LR \newline NN \newline RF & \textbf{51.3} {\scriptsize$\pm$ 1.0}\newline  \textbf{49.9} {\scriptsize$\pm$ 1.3}\newline \textbf{50.0} {\scriptsize$\pm$ 0.0} & 62.3 {\scriptsize$\pm$ 0.8}\newline 99.0 {\scriptsize$\pm$ 0.1}\newline \textbf{61.1} {\scriptsize$\pm$ 0.4} & 96.0 {\scriptsize$\pm$ 0.2}\newline 96.7 {\scriptsize$\pm$ 0.5}\newline 96.4 {\scriptsize$\pm$ 0.2}\\
	
	\end{tabu}
	\label{tab:fb}
\end{table}

\begin{table}
    \caption{Representation Learning on Cora Dataset}
	%\fontsize{9pt}
	%\setlength{\arraystretch}{0.5}
	\linespread{1.0} 
	\setlength{\tabcolsep}{3pt} % Default value: 6pt
	\centering
	\begin{tabu}{l|p{11mm}|p{13mm}|p{13mm}|p{13mm}}
	
	Method & Classifier & Node RB $\downarrow$ & Link RB $\downarrow$ & Accuracy $\uparrow$\\
    \midrule
    DeBayes & LR \newline NN \newline RF & \textbf{21.7} {\scriptsize$\pm$ 1.8} \newline  66.7 {\scriptsize$\pm$ 1.6}\newline 25.8 {\scriptsize$\pm$ 1.2}  & \textbf{26.7} {\scriptsize$\pm$ 0.9}\newline \textbf{75.6} {\scriptsize$\pm$ 2.5}\newline 35.9 {\scriptsize$\pm$ 0.2} & 72.8 {\scriptsize$\pm$ 1.9}\newline 78.5 {\scriptsize$\pm$ 1.6} \newline 76.9 {\scriptsize$\pm$ 1.3}\\
    \midrule
    FairWalk & LR \newline NN \newline RF & 67.1 {\scriptsize$\pm$ 0.7}\newline  71.9 {\scriptsize$\pm$ 1.9}\newline 35.2 {\scriptsize$\pm$ 1.5} & 80.3 {\scriptsize$\pm$ 0.5} \newline 86.9 {\scriptsize$\pm$ 0.8} \newline 49.2 {\scriptsize$\pm$ 1.5} & \textbf{83.0} {\scriptsize$\pm$ 0.7} \newline \textbf{85.6}
    {\scriptsize$\pm$ 0.6}\newline \textbf{80.3} {\scriptsize$\pm$ 0.2}\\
    \midrule
    Node2Vec+FairDrop & LR \newline NN \newline RF & 56.0 {\scriptsize$\pm$ 1.0}\newline \textbf{58.5} {\scriptsize$\pm$ 2.3}\newline \textbf{19.4} {\scriptsize$\pm$ 0.1} & 73.8 {\scriptsize$\pm$ 1.2}\newline 82.0 {\scriptsize$\pm$ 0.6}\newline \textbf{32.1} {\scriptsize$\pm$ 1.4} & 78.0 {\scriptsize$\pm$ 0.8}\newline 76.7 {\scriptsize$\pm$ 1.0}\newline 75.4 {\scriptsize$\pm$ 1.2}\\
	
	\end{tabu}
	\label{tab:cora_emb}
\end{table}

\begin{table}
    \caption{Representation Learning on Citeseer Dataset}
	%\fontsize{9pt}
	%\setlength{\arraystretch}{0.5}
	\linespread{1.0} 
	\setlength{\tabcolsep}{3pt} % Default value: 6pt
	\centering
	\begin{tabu}{l|p{11mm}|p{13mm}|p{13mm}|p{13mm}}
	
	Method & Classifier & Node RB $\downarrow$ & Link RB $\downarrow$ & Accuracy $\uparrow$\\
    \midrule
    DeBayes & LR \newline NN \newline RF & \textbf{24.8} {\scriptsize$\pm$ 1.7} \newline  56.7 {\scriptsize$\pm$ 1.0}\newline 35.4 {\scriptsize$\pm$ 2.0}  & \textbf{28.9} {\scriptsize$\pm$ 2.1}\newline \textbf{50.7} {\scriptsize$\pm$ 1.5}\newline 40.3 {\scriptsize$\pm$ 2.6} & 70.0 {\scriptsize$\pm$ 1.7}\newline 78.5 {\scriptsize$\pm$ 0.9} \newline 74.1 {\scriptsize$\pm$ 1.2}\\
    \midrule
    FairWalk & LR \newline NN \newline RF & 49.1 {\scriptsize$\pm$ 1.5}\newline  59.1 {\scriptsize$\pm$ 1.3}\newline 44.1 {\scriptsize$\pm$ 1.7} & 56.1 {\scriptsize$\pm$ 0.3} \newline 63.9 {\scriptsize$\pm$ 1.5} \newline 44.2 {\scriptsize$\pm$ 0.8} & \textbf{79.7} {\scriptsize$\pm$ 0.7} \newline \textbf{82.0} {\scriptsize$\pm$ 0.7}\newline \textbf{74.3} {\scriptsize$\pm$ 0.5}\\
    \midrule
    Node2Vec+FairDrop & LR \newline NN \newline RF & 39.4 {\scriptsize$\pm$ 0.4}\newline  \textbf{39.7} {\scriptsize$\pm$ 0.8}\newline \textbf{31.6} {\scriptsize$\pm$ 0.7} & 54.5 {\scriptsize$\pm$ 1.0}\newline 61.0 {\scriptsize$\pm$ 0.4}\newline \textbf{38.1} {\scriptsize$\pm$ 1.0} & 75.8 {\scriptsize$\pm$ 0.6}\newline 76.9 {\scriptsize$\pm$ 0.7}\newline 74.1 {\scriptsize$\pm$ 0.5}\\
	
	\end{tabu}
	\label{tab:citeseer_emb}
\end{table}

\begin{table}
    \caption{Representation Learning on PubMed Dataset}
	%\fontsize{9pt}
	%\setlength{\arraystretch}{0.5}
	\linespread{1.0} 
	\setlength{\tabcolsep}{3pt} % Default value: 6pt
	\centering
	
	\begin{tabu}{l|p{11mm}|p{13mm}|p{13mm}|p{13mm}}
	
	Method & Classifier & Node RB $\downarrow$ & Link RB $\downarrow$ & Accuracy $\uparrow$\\
    \midrule
    DeBayes & LR \newline NN \newline RF & \textbf{49.3} {\scriptsize$\pm$ 0.3} \newline  67.7 {\scriptsize$\pm$ 1.2}\newline 49.4 {\scriptsize$\pm$ 2.0}  & \textbf{55.5} {\scriptsize$\pm$ 1.9}\newline 83.6 {\scriptsize$\pm$ 0.7}\newline 57.7 {\scriptsize$\pm$ 0.6} & 80.3 {\scriptsize$\pm$ 0.3}\newline 82.2 {\scriptsize$\pm$ 0.4} \newline \textbf{80.6} {\scriptsize$\pm$ 0.5}\\
    \midrule
    FairWalk & LR \newline NN \newline RF & 68.1 {\scriptsize$\pm$ 0.3}\newline  67.5 {\scriptsize$\pm$ 0.8}\newline 53.2 {\scriptsize$\pm$ 0.8} & 74.5 {\scriptsize$\pm$ 0.7} \newline 83.7 {\scriptsize$\pm$ 0.3} \newline 53.2 {\scriptsize$\pm$ 0.8} & \textbf{82.1} {\scriptsize$\pm$ 0.2} \newline \textbf{88.1} {\scriptsize$\pm$ 0.3}\newline 76.2 {\scriptsize$\pm$ 0.6}\\
    \midrule
    Node2Vec+FairDrop & LR \newline NN \newline RF & 56.2 {\scriptsize$\pm$ 0.3}\newline  \textbf{54.1} {\scriptsize$\pm$ 1.0}\newline \textbf{44.6} {\scriptsize$\pm$ 0.6} & 70.5 {\scriptsize$\pm$ 0.4}\newline \textbf{80.2} {\scriptsize$\pm$ 0.3}\newline \textbf{55.6} {\scriptsize$\pm$ 0.6} & 79.8 {\scriptsize$\pm$ 0.6}\newline 79.3 {\scriptsize$\pm$ 0.5}\newline 76.3 {\scriptsize$\pm$ 0.8}\\
	
	\end{tabu}
	\label{tab:pubmed_emb}
\end{table}

Node2Vec coupled with FairDrop improves the obfuscation of the sensitive attribute in the resulting node embeddings with the biggest gain in the Node RB evaluation. Furthermore, the obfuscation is robust against the choice of the downstream classifier. This improvement comes with a small drop in the accuracy of the downstream task.
The Link RB scores are higher than the previous ones due to the additional information provided by the concatenation of the two nodes embeddings forming the link. In this case, FairDrop performs on par with DeBayes if we limit our observation to the number of times the algorithm got the best score.
FairWalk is less aggressive,  resulting in less protection but higher accuracy scores on the downstream tasks.
Pairing Node2Vec with FairDrop is orders of magnitudes faster than DeBayes. It takes just 50 seconds, or 1 second per epoch, against  33 minutes on average required by DeBayes for converging in its standard configuration. FairDrop and FairWalk share a similar approach in tackling the fairness issue. However, FairWalk needs to evaluate the neighbourhood of each node in the random walk during its construction. This implementation is harder to optimize, and it requires a fixed cost of almost 80 seconds to build the random walks and further 200 seconds to train the learning module. All these quantities refer to the DBLP dataset.

\subsection{Ablation}
\label{sec:ablation}
\begin{figure}
    \centering
    \includegraphics[width=\columnwidth]{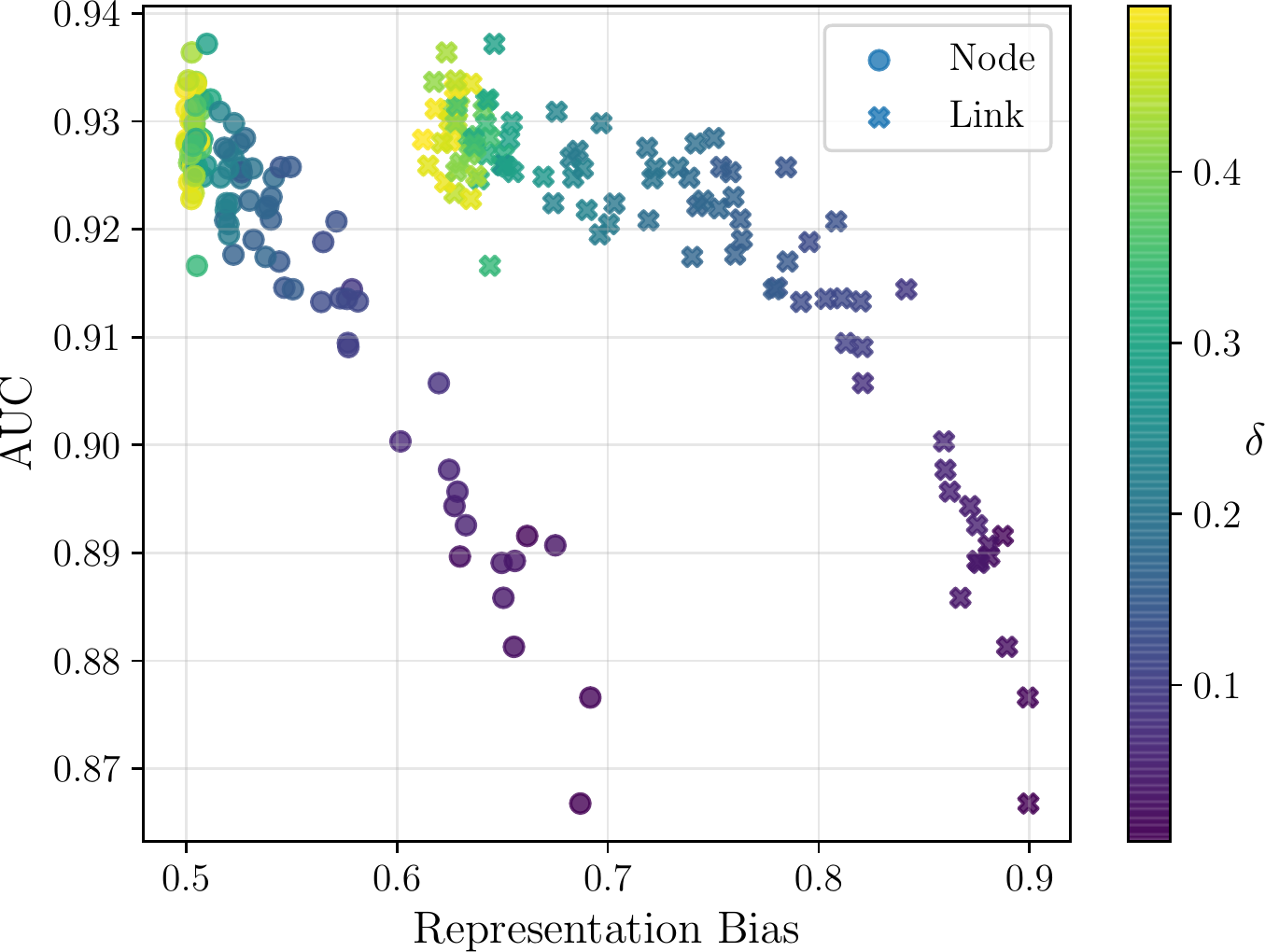}
    \caption{Visualization on the FB dataset of the trade-off between the representation biases and the performances on the downstream link prediction task.}
    \label{img:ablation}
\end{figure}

\begin{figure}
    \centering
    \includegraphics[width=\columnwidth]{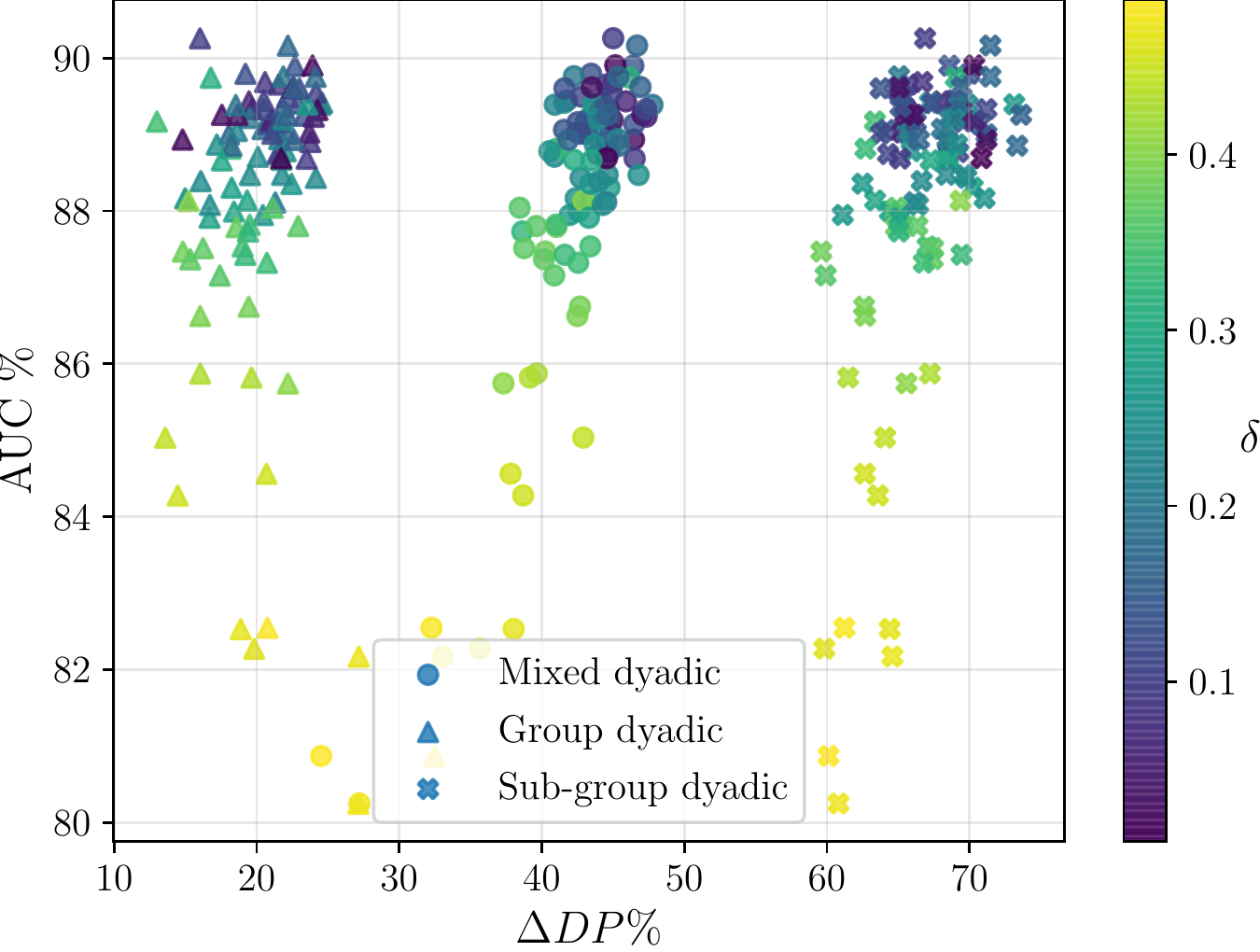}
    \caption{Visualization on the Citeseer dataset of the trade-off between the demographic parity difference across the dyadic groups and the downstream link prediction task.}
    \label{img:link_ablation}
\end{figure}

We perform an ablation study on the behaviour imposed by the parameter $\delta$. We report in Figure \ref{img:ablation} the AUC score and both representation biases obtained using a random forest classifier on a single data split for one hundred runs on the FB dataset. We recall that for a higher value of $\delta$, the sampling is more biased towards fairness. Surprisingly the AUC of the downstream link prediction task grows as the representation biases decrease. Despite seeming counter-intuitive, we think this is in line with what the authors of \cite{chen2020nodehomo} observed. They examined the problem of over-smoothing of GNNs for node classification tasks. They argued that one factor leading to this issue is the over-mixing of information and noise. In particular intra-class connections can bring valuable information, while inter-class interaction may lead to indistinguishable representations across classes. 
However, for the link prediction task, the inverse is true. An excessive similarity between intra-class embedding and dissimilarity between inter-class ones results in the known ``filter bubble" issue. In Figure \ref{img:link_ablation} we repeat the same ablation study for the end-to-end link prediction task. The plot shows the trade-off between the $\Delta DP$ across the different dyadic groups and the AUC score of the link prediction. For $\delta \in (0.2,0.3)$, we have better or equal AUC scores paired with a fairness improvement. Over a certain value of $\delta$, the AUC score has a substantial drop. A strongly biased dropout drastically changes the topology removing some, or all, the randomness of the input and its beneficial effects. Considering the small size of the datasets and the 2-hop aggregation mechanism, these are the potential causes for the issue.

\section{Conclusion}
\label{sec:conclusion}
We proposed a flexible biased edge dropout algorithm for enhancing fairness in graph representation learning. FairDrop targets the negative effect of the network's homophily with respect to the sensitive attribute.
A single hyperparameter regulates the intensity of the constraint, ranging from maximum fairness to an unbiased edge dropout. We tested FairDrop performances on two common benchmark tasks, specifically unsupervised node embedding generation and link prediction. It is possible to frame FairDrop as a biased data augmentation pipeline. Therefore it can be used with different architectures and paired with other fairness constraints.

To measure the gains, we adapted the group fairness metrics to estimate the disparities between the dyadic groups defined by the edges. We proposed a new dyadic group aimed to be closer to the original groups defined by the sensitive attributes. Finally we extended the previous metric evaluating the representation bias to take into account the graph structure. We plan to develop these metrics for the case of a strong imbalance of the sensitive attributes.

A possible limitation of FairDrop is its extension to the case where multiple sensitive are present at once. We proposed to alternate the masks of the different sensitive attributes to guide the biased dropout. However, this pipeline may be sub-optimal.

Randomness is a fundamental concept for FairDrop. However, not every connection has the same importance inside a graph. Future lines of research may combine perturbation-aware algorithms \cite{ceci2020} with fairness constraints to perform a more selective edge sampling procedure.

\section{Acknowledgements}
The authors are grateful to the anonymous reviewers for their insightful comments and suggestions. Hussain would like to acknowledge the support of the UK Engineering and Physical Sciences Research Council (EPSRC) - Grants Ref. EP/M026981/1, EP/T021063/1, EP/T024917/1.

\bibliographystyle{ieeetr}
\bibliography{biblio}
\end{document}